\documentclass[nohyperref]{article}

\usepackage{url}
\usepackage[textsize=small]{todonotes}

\usepackage{microtype}
\usepackage{graphicx}
\usepackage{booktabs} 
\usepackage{subcaption}
\usepackage{multirow}
\usepackage{wrapfig}

\usepackage[accepted]{icml2022}
\usepackage{hyperref}

\usepackage{icml2022}

\usepackage{amsmath}
\usepackage{amssymb}
\usepackage{mathtools}
\usepackage{amsthm}

\usepackage[capitalize,noabbrev]{cleveref}

\theoremstyle{plain}

\theoremstyle{definition}

\theoremstyle{remark}

\icmltitlerunning{Data Scaling Laws in NMT: The Effect of Noise and Architecture}

\begin{document}

\twocolumn[
\icmltitle{Data Scaling Laws in NMT: The Effect of Noise and Architecture}




\begin{icmlauthorlist}
\icmlauthor{Yamini Bansal}{yyy}
\icmlauthor{Behrooz Ghorbani}{comp}
\icmlauthor{Ankush Garg}{comp}
\icmlauthor{Biao Zhang}{sch}
\icmlauthor{Maxim Krikun}{comp}
\icmlauthor{Colin Cherry}{comp}
\icmlauthor{Behnam Neyshabur}{comp}
\icmlauthor{Orhan Firat}{comp}
\end{icmlauthorlist}

\icmlaffiliation{yyy}{School of Engineering and Applied Science, Harvard University, MA, USA. Work performed while interning at Google.}

\icmlaffiliation{comp}{Google, USA}
\icmlaffiliation{sch}{School of Informatics, University of Edinburgh}

\icmlcorrespondingauthor{Yamini Bansal}{ybansal@g.harvard.edu}

\icmlkeywords{Machine Learning, ICML}

\vskip 0.3in
]



\printAffiliationsAndNotice

\begin{abstract}
In this work, we study the effect of varying the architecture and training data quality on the data scaling properties of Neural Machine Translation (NMT). First, we establish that the test loss of encoder-decoder transformer models scales as a power law in the number of training samples, with a dependence on the model size. Then, we systematically vary aspects of the training setup to understand how they impact the data scaling laws. In particular, we change the following (1) Architecture and task setup: We compare to a transformer-LSTM hybrid, and a decoder-only transformer with a language modeling loss (2) Noise level in the training distribution: We experiment with filtering, and adding iid synthetic noise. In all the above cases, we find that the data scaling exponents are minimally impacted, suggesting that marginally worse architectures or training data can be compensated for by adding \emph{more data}. Lastly, we find that using back-translated data instead of parallel data, can significantly degrade the scaling exponent.

\end{abstract}

\section{Introduction}

Scaling up the amount of data used for training has emerged as a robust way to make progress on various tasks in deep learning \citep{krizhevsky2012imagenet, brown2020language, kolesnikov2020big, thoppilan2022lamda}. 
On the other hand, research on new architectures and data filtering methods that beat fixed dataset benchmarks continues to grow. However, it is unclear how many of these improvements in the architecture \& data quality translate to improvements in the \emph{sample efficiency}, or the rate with which these models learn from increasing amounts of data.

Recent work \citep{Hestness2017DeepLS, rosenfeld2019constructive, kaplan2020scaling} on \emph{scaling laws} offers a useful tool to answer this question --- they show that the test loss of a model scales predictably as a power law in the relevant quantities of interest such as dataset size ($D$). Thus, the sample efficiency of a training algorithm over many orders of magnitudes of data can be captured succinctly with a scaling law.

We take these findings a step further --- we conduct a large-scale study in Neural Machine Translation (NMT) to understand \emph{how different interventions to the training setup impact the data scaling laws} (See Figure \ref{fig:intro} for a summary of our results). In particular, we take the most common and effective tools used to improve performance, architecture and sources/quality of data, and investigate how changing them affects the data scaling law. In particular, we make the following contributions:

\begin{figure*}[t]
    \centering
    \includegraphics[width=\linewidth]{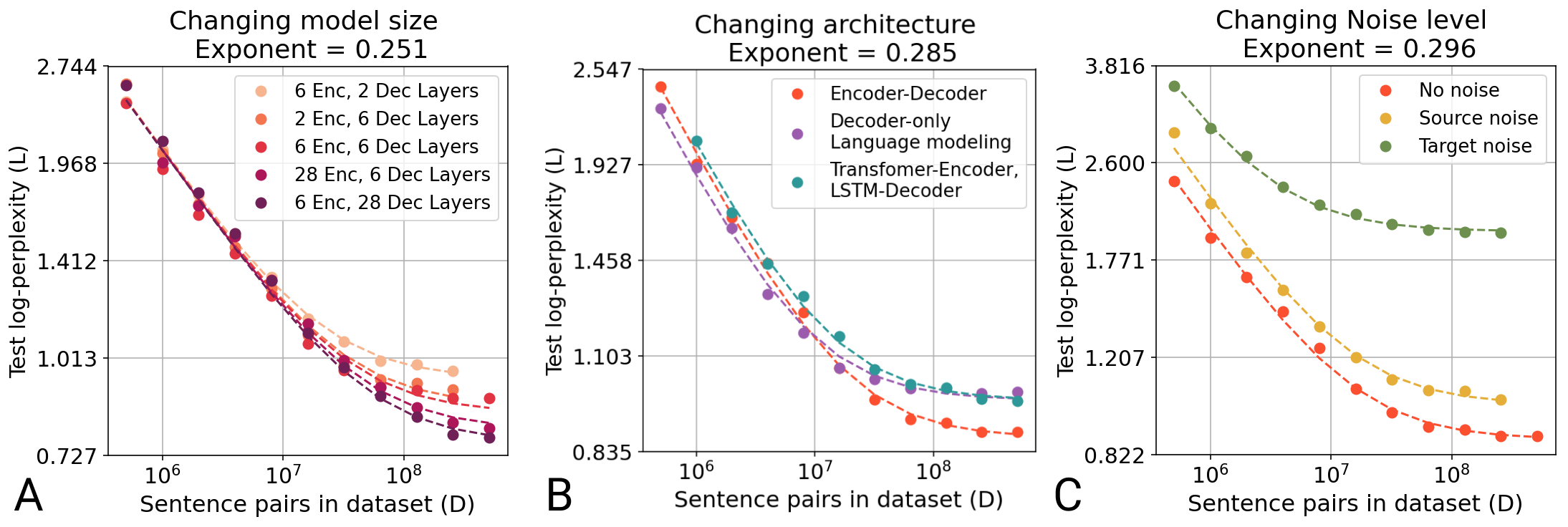}
    \caption{\textbf{Data scaling exponent is minimally impacted by changes to the training setup.} We train a series of models on increasing amounts of data $\{500K...512M\}$ sentence pairs while (A) Changing the depth \& the shape of the encoder-decoder transformer (B) Changing the architecture and task setup (C) Changing the noise in the training distribution. For each figure, we fit the data scaling law similar Eq. \eqref{eg:basic-scaling-law} and find that a common exponent $p$ provides good fits for the empirical observations. See sections \ref{sec:basic-scaling}, \ref{sec:architecture} \& \ref{sec:noise} for full details.}
    \label{fig:intro}
\end{figure*}

{\bf Our Contributions:} We first establish that the test log-perplexity of encoder-decoder transformer models trained on an English $\rightarrow$ German translation task evolves as a smooth function of the dataset size, with a dependence on the model size (Figure \ref{fig:intro}A). We demonstrate that our scaling law predicts experimental results over $3$ orders of magnitude of training examples ( (from 500K-512M sentence pairs) \footnote{Corresponding to 27.6 billion tokens.}. Then, we systematically vary the following aspects of the training setup to understand how they impact scaling laws: 

\textit{\textbf{Architecture:}} We compare encoder-decoder transformers of different shapes, transformer-LSTM hybrids and decoder-only transformers. We also verify our results for a different language pair (Chinese $\rightarrow$ English). \\ 
\textit{\textbf{Data Source and Filtering:}} We consider two different data sources with different data crawling approaches. We also study the role of data filtering on sample efficiency by examining two different filtering algorithms, Bicleaner \cite{ramirez-sanchez-etal-2020-bifixer} and Contrastive Data Selection (CDS) \citep{wang2018denoising}.\\
\textit{\textbf{Synthetic Noise:}} We add synthetic iid noise to the source (input) and the target (output) side separately. \\
\textit{\textbf{Back-translation:}} We train with back-translated (BT) \cite{sennrich2016improving} data with varying BT model sizes.

In all, we produce 20 different data scaling curves, each consisting of 10 different dataset sizes which take $\approx 25K$ TPUv3-hours to produce.


We find that, with the exception of back-translation, these changes \emph{do not impact the scaling exponents significantly} (See Fig. \ref{fig:intro} for examples). Thus, our work shows that many of the common operations used to boost performance, such as small changes to the architecture or data filtering, are mere scaling penalties. That is, in some cases \emph{\textbf{sub-optimalities in the architectures and data quality can be compensated for by adding an extra constant factor of data}}. \footnote{We would like to qualify that this statement is based on empirical evidence and is not expected to hold with full generality. We do not understand yet if this finding would hold for very different classes of architectures or noise.}


The utility of our detailed study is multi-fold. Practically for NMT, data scaling laws can be leveraged to make experimental decisions for future large-scale experiments. 
This is especially important because current NMT models are trained using massive web-scale data; \citet{arivazhagan2019massively} used 25B training examples (approx.1T tokens) and with the advent of the self-supervised learning techniques \citep{liu2020multilingual,raffel2020exploring,siddhant2020leveraging} this number can easily reach 10+T tokens. At such large scales of data, it is unfeasible to `just perform the experiment', and scaling laws can be used to drive training decisions. For instance, if small changes in architecture do not lead to a change in scaling exponent (as shown in Figure \ref{fig:intro}B), then architecture choice can be driven by other factors such as computational efficiency in exchange for a small penalty of more data. Our findings are also theoretically significant --- since the scaling exponent can be interpreted as a `signature' of the underlying learning mechanism (See \citet{sharma2020neural, bahri2021explaining} for possible hypothesis), our results show that seemingly different training methods have deeper commonalities worthy of further investigation.

\subsection{Experimental Setup}
\label{sec:exp-setup}

\textbf{Models:} Our experiments are conducted on pre-layer transformer networks \citep{xiong2020layer}. Models are trained with per-token cross-entropy loss and Adafactor optimizer \citep{shazeer2018adafactor}. All models are trained with a fixed batch-size of 500K tokens and dropout rate of $0.1$ for residuals, feed-forward activations and attention. For the small dataset sizes, the models are trained to early stopping (as measured on the log-perplexity of a held-out development set) and for large dataset sizes they are trained for up to 500K gradient steps. The hyperparameters for these models were optimized for a 6 encoder layer and 6 decoder layer model trained on 2.2 billion sentence pairs. For a detailed discussion of the effect of hyperparameters, please see Appendix \ref{app:hyperparams}. 

We also train two decoder-only models with a language modeling loss with \{9L, 13L\}, and three hybrid-LSTM model with \{6L2L, 6L6L, 6L12L\}. All the hyperparameters are matched as closely as possible between these models to provide an apples-to-apples comparison.

\textbf{Training Data} In our experiments, the models are trained on two large-scale datasets. The first set of experiments are conducted using an in-house parallel corpora containing up to 2.2B sentences translated from English to German. We sample training datasets of sizes \{1M, 2M, 4M, 8M, 16M, 32M, 64M, 128M, 256M, 512M\} independently to study the data scaling laws. The second set of experiments are conducted with Paracrawl dataset \citep{banon-etal-2020-paracrawl}, both with and without filtering applied. The details are described in Section \ref{sec:noise}. To verify that our results generalize to different language pairs, we repeat a subset of our experiments on an in-house Chinese to English dataset with a similar training setup. 

\textbf{Test Data} The model performance is measured on a held-out dataset from the training distribution. We also measure performance on various out-of-distribution test datasets that have different domain composition and sampling methods. Please refer to Section \ref{sec:ood} for a detailed description. 

\section{Related Works}
Our work builds extensively on the literature on scaling laws \citet{amari, Hestness2017DeepLS, rosenfeld2019constructive} and in particular \citet{kaplan2020scaling}. 

Prior works that have studied scaling laws in NMT include \citet{ghorbani2021scaling, Hestness2017DeepLS, gordon2021data}. Our experimental setup shares various commonalities with \citet{ghorbani2021scaling} but with the important distinction that Ghorbani et al. study scaling with respect to the \emph{number of parameters}, while we study scaling with respect to the dataset size. Thus, these works should be considered complementary --- they drive experimental decision making in different regimes and can give qualitatively different recommendations as we describe in Section \ref{sec:basic-scaling}. 

\citet{gordon2021data} also consider data scaling laws for NMT. Our work differs from them in the following fundamental ways (1) While they focus on establishing data and parameter scaling laws for NMT, our main goal is to examine the role of design choices such as model architecture and data collection methods from the perspective of scaling laws
(2) Gordon et al. focus on the small data/small model regime (maximum of 50M sentence pairs). In contrast, we focus on the performance of models at web-scale data (ranging up to 1B sentence pairs). Working in the smaller data regime allows \citet{gordon2021data} to make better predictions about the behavior of NMT systems for low-resource languages, while working in the large data regime allows us to make better predictions for high-resource languages and assess the effectiveness of interventions such as filtering (where having a large dataset allows us to throw away data and still observe reasonable scaling laws). 

Our scaling law differs from \cite{Hestness2017DeepLS} in that they conduct experiments with LSTMs and their law does not scale as $O(1/D)$ when $D\rightarrow\infty$. Despite these differences, the scaling exponents found in both these papers are in the same range as ours $0.25-0.3$. Note that these exponents are much higher than those found for the unconditional language modeling case by \citet{kaplan2020scaling, Hestness2017DeepLS} ($0.1$ vs. $0.3$).
In the vision domain, \citet{pmlr-v139-hoiem21a} compare data scaling laws of different architectures and pre-training methods. \cite{param_scaling_transformers} also studies the inductive bias of architectures using scaling, but they compare parameter scaling, while we compare data scaling.

\section{Data Scaling Laws}
\label{sec:basic-scaling}

We begin our investigation by training a series of large-scale encoder-decoder transformer models on an in-house English to German parallel corpus, with the parameters ranging from 170M to 800M. Our dataset sizes range from 500K to 512M sentence pairs, covering 28B tokens. We train $5$ different model sizes \{2L6L, 6L2L, 6L6L, 6L28L, 28L6L\}, where 28L6L means 28 encoder layers and 6 decoder layers. We will mainly focus on the test log-perplexity on a heldout dataset from the training distribution, but we will also discuss scaling of BLEU scores (Section \ref{sec:bleu}) and performance on out-of-distribution test sets (Section \ref{sec:ood}). 


\begin{figure*}
    \centering
    \includegraphics[width=\linewidth]{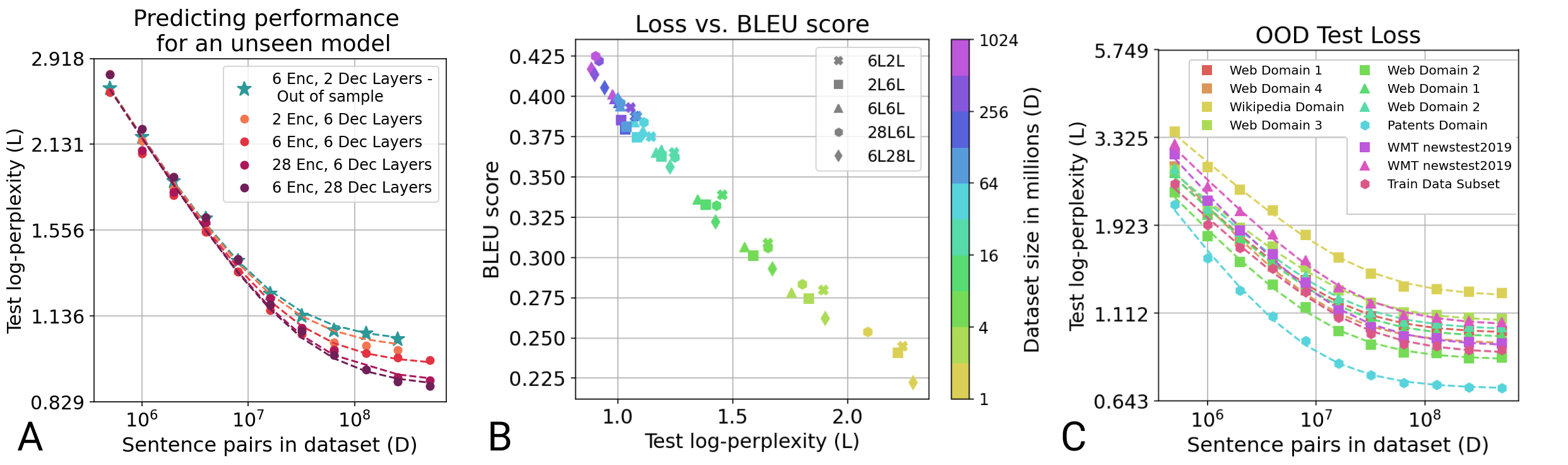}
    \caption{(A) Using the joint data and model scaling law from Eq. \eqref{eg:joint-scaling-law} to predict the performance of a previously unseen model. Results are plotted for Web Domain 1 test set (target original). (B) BLEU score and test loss display an almost-linear relationship (C) Scaling law for different OOD test sets for a 6L6L model have similar exponents. The square and triangle markers show the source-original and target-original test sets respectively.}
    \label{fig:joint-oos}
\end{figure*}
{\bf Form of the Scaling Law:} The chosen scaling law must exhibit decreasing loss with dataset size $D$ (measured in millions of sentence pairs). At infinite data $D=\infty$, the model must converge to a finite constant. Additionally, \citet{kaplan2020scaling, bahri2021explaining}, conjecture that when $D \rightarrow \infty$, the models are in a ``variance-limited" regime and the loss should scale as $O(1/D)$. These desiderata are satisfied by the following scaling law for a fixed model size:

\begin{equation}
    L(D) = \alpha\Big(D^{-1} + C\Big)^p
    \label{eg:basic-scaling-law}
\end{equation}

where $\alpha$ is a multiplicative constant, $p$ is the scaling exponent and $C$ is a constant corresponding to the model capacity. We find that this scaling law indeed provides a good fit for the experimental observations. Moreover, when we fit just the last few datapoints (See Figure \ref{fig:var-limited}), we find that the loss scales as $O(1/D)$, thus confirming variance-limited scaling.

Another important requirement for any scaling law in an encoder-decoder setting is that it should be robust to the variations in the network shape. In NMT, many practitioners often use encoder-heavy models due to their inference efficiency \citep{kasai2020deep}. In contrast, decoder-heavy models have shown great promise in applications such as conversational AI \citep{xu2020recipes}. Prior work \citep{ghorbani2021scaling} has demonstrated that, in the context of parameter scaling, encoder-heavy models exhibit significantly different behaviors compared to decoder-heavy models. 

To examine how model shape interacts with sample efficiency, we examine the data scaling characteristics of $5$ different models with various degrees of encoder-decoder asymmetry: \{2L6L, 6L2L, 6L6L, 28L6L, 6L28L\}. Our results indicate that the same scaling parameters $\alpha, p$ are sufficient to capture the data scaling behavior for all of these models; the only parameter that has to change from model to model is $C$ which is due to the difference in the model capacities. See Figure \ref{fig:intro}A for the scaling law fit and Appendix \ref{app:separate-fits} for individual fits of Eq. \eqref{eg:basic-scaling-law} for each model. As such, our results indicate that \emph{data efficiency characteristics of encoder-decoder NMT models is largely independent of the model shape}.

{\bf Joint Data \& Parameter Scaling Law:} In Eq. \eqref{eg:basic-scaling-law}, as $D \rightarrow\infty$, $L(D) \rightarrow \alpha C^p$ which corresponds to the achievable test loss when the learning problem is limited only by the model capacity. Earlier work has studied the behavior of $\alpha C^p$ in the context of parameter scaling laws \citep{ghorbani2021scaling, gordon2021data}. Here, we leverage these results to derive a joint data \& parameter scaling law. To be specific, instead of empirically fitting the constant $C$, we replace it with the quantity implied by the model scaling law computed by Ghorbani et al.: 
\begin{align}    \label{eg:joint-scaling-law}
    \begin{split}
    L(D; N_e, N_d) &= \alpha\Big(D^{-1} + C_{N_e, N_d} \Big)^p,  \\
    C_{N_e, N_d} &= \beta\Big( N_e^{-p_e}N_d^{-p_d} + L_{\infty} \Big)^{1/p}. 
    \end{split}
\end{align}
Here, $N_e$ and $N_d$ correspond to the number of parameters in the encoder \& decoder respectively. The only fitted parameters in this equation are $\alpha$ and $p$; the parameters $(\beta, p_e, p_d, L_{\infty})$ appearing in \eqref{eg:joint-scaling-law} are directly borrowed from \citet{ghorbani2021scaling}. As such, $L(D; N_e, N_d)$  converges to the parameter scaling law as $D\rightarrow \infty$. Figure \ref{fig:joint-oos}A shows that the joint law is able to closely capture the combined effects of data \& model size variation across all our models. 

We also examine the performance of our joint law in predicting the test loss of data-model combinations that were not used in fitting the scaling law. Specifically, we fit $\alpha, p$ using the empirical test loss values of all data-model combinations except the ones using a 6L2L model. Then we examine how well our joint law is able to predict the test performance of the held-out models. Figure \ref{fig:joint-oos}A shows both the empirical \& the predicted test loss values. As the figure suggests, our joint law is able to accurately predict the test performance of models \emph{out-of-sample}.


As a final sanity check, we examine the robustness of our scaling laws to variance arising from randomness in training (See Appendix \ref{app:separate-fits}) and from variation in hyperparameters (See Appendix \ref{app:hyperparams}). 


{\bf Implications:} Eq. \eqref{eg:basic-scaling-law} suggests that there exists two operating regimes for data scaling: (i) data-limited regime where $D^{-1} \gg C$, and (ii) capacity limited regime where $D^{-1} \ll C$. Fitted exponents in Figure \ref{fig:intro} suggest that, in the data limited regime, loss scales as $O(D^{-1/4})$, suggesting a marginal value of $O(D^{-5/4})$ for additional data. Increasing the model capacity in this regime has negligible impact on the loss. In the capacity limited regime however, the loss scales as $O(D^{-1})$ suggesting a (significantly smaller) marginal value of $O(D^{-2})$ for additional data. In this regime, the loss value is dominated by the model-dependent constant $C$ and most of the improvement can be had by increasing the model size. There is a smooth \emph{phase transition} between these two regimes at approximately $CD = 1$. See Appendix \ref{app:phase-transition} for an illustration. Thus, by increasing the model size (which reduces $C$), one can push the transition to larger values of $D$ and leverage the available data more efficiently. 



\subsection{BLEU Score}
\label{sec:bleu}
Machine translation is both a language \emph{understanding} as well as a language \emph{generation} task: source content first needs to be understood and the corresponding target sequence must be generated. Given this categorization, we not only care about the model score on reference target sequence (measured in log-perplexity) but we also care about the generation quality. While evaluation of generation quality is an active research area, we compute automatic measures like BLEU score as a proxy for the quality of the generated targeted sentences. See Appendix \ref{app:bleu} for details on BLEU score calculation and an extended discussion on the challenges of evaluating generation quality. We find that the BLEU score and test log-perplexity have a nearly linear relationship as shown in Figure \ref{fig:joint-oos}B. Thus, the BLEU score also scales predictably with the dataset size.

\subsection{Out-of-Distribution Generalization}
\label{sec:ood}
To get a more robust understanding of the generalization capabilities of our models, we evaluate them on a collection of out-of-distribution (OOD) evaluation sets. These test sets cover a number of different domains: (i) Web-Domain (ii) News-Domain (iii) Patents (iv) Wikipedia. While the majority of our test sets are internal, News-Domain test sets come from WMT2019 \citep{barrault-EtAl:2019:WMT} evaluation campaign (newstest2019). In addition, these test sets are constructed from a diverse set of composition approaches: Most of the test sets are source-original, i.e., sentence pairs are formed by translating natural human text from the source language to the target language. However, for some domains, we also have target-original test sets where natural target sentences are backwards translated to the source sentence. Earlier research has demonstrated the importance of differentiating between the two composition approaches as the style of natural sentences and translated sentences is different \citep{Freitag19, freitag-etal-2020-bleu, graham2020statistical}.

We find that the test loss on these test sets also follows a scaling law described by Eqn. \eqref{eg:basic-scaling-law}. Figure \ref{fig:joint-oos}C shows the test loss fits for all the test sets for a 6L6L model. See Figure \ref{fig:lang-ood} for similar results for Chinese to English. Previous research \citep{ghorbani2021scaling} has demonstrated that, when scaling the model size, source and target original test sets exhibit drastically different generalization dynamics. Surprisingly, in the data scaling context, we find that most of the test sets have similar scaling exponents (See Figure \ref{fig:app-diff-tests}C for exact values). That is, we do not observe any major systematic differences in data scaling behavior on the basis on the test set composition --- both target-original and source-original test sets scale similarly. 

Additionally, since the different test sets have similar scaling exponents, which implies that the in-distribution loss and the out-of-distribution loss have a nearly-linear relationship (See Figure \ref{fig:app-diff-tests}C). This is in line with previous findings in vision \citep{miller2021accuracy}. Why these distributions scale similarly (or why they have a linear relationship) is still an open research question.

\section{The Effect of Architecture}
\label{sec:architecture}


The model architecture is a key tool in the machine learning repertoire to improve performance. However, most evaluations of architectures are performed for a fixed dataset size. Instead, in this section, we consider how the performance of an architecture scales as the dataset size increases.

{\bf Setup:} We pick three architecture and loss setups that are commonly used for machine translation. We take an encoder-decoder transformer discussed in Section \ref{sec:basic-scaling} as the baseline. Next, we pick a hybrid architecture with a transformer encoder and an LSTM decoder \citep{chen2018best} due to their wide adoption by the industry applications \citep{MSFTHybrid, GoogleHybrid}.
This allows us to compare the sample efficiency of a transformer vs. an LSTM --- both popular but different sequence-to-sequence architectures. Finally, we use a decoder-only transformer that is trained with a language modeling (LM) loss. Thus, the last model changes not only the architecture, but also the loss (by including an LM loss on the source side). This setup mimics the GPT series of models, and has also been shown to perform well in MT \cite{wang2021language}. We train the models on increasing subsets of the data as described in Section \ref{sec:exp-setup}. All the three models have $\sim 300M$ parameters. We also compare encoder-decoder transformers and hybrid models for Chinese to English translation in Appendix \ref{app:diff-lang}, showing that our results are independent of the choice of language pair. 

{\bf Results:} In Figure \ref{fig:intro}B, we fit a scaling law with a common exponent $p$, but model-dependent $\alpha, C$ to the data. As the figure suggests, we find that a scaling law with a common exponent closely captures the behavior of the observed experimental data. \footnote{ See Appendix \ref{app:architecture} for more results with different model sizes of the hybrids and decoder-only models.} Our experiments show that while different architectures may have different performances on a fixed dataset size, the architectures may nevertheless scale similarly with more data in the data-limited regime. Thus, \textit{\textbf{we can compensate for a marginally worse architecture by adding more data}}. Crucially, the factor of additional data to be added does not depend on the loss value --- it will always be $\frac{\alpha_1}{\alpha_2}^{1/p}$, where $\alpha_1, \alpha_2$ are the multiplicative constants for the two architectures and $p$ is the scaling exponent. If the exponents were different, the amount of data to achieve equal performance would \emph{increase exponentially} with decreasing loss. This suggests that when choosing between multiple architectures that have similar data scaling, the decision can freely be driven by other considerations such as compute efficiency, multi-task abilities, compressibility for deployment or generation latency. Moreover, this suggests that minor tweaks to architectures will only improve the scaling constant and not the exponent. So, efforts can be redirected towards gathering more data to obtain the same performance gains.

\begin{table*}[t]
\caption{{\bf Scaling coefficients:} In all the settings, we fit a scaling law of the form $\alpha_{model}(D^{-1} + C_{model})^p$. Note the common exponent $p$.}
\vskip 0.15in
\begin{center}
\begin{small}
\begin{sc}
\begin{tabular}{|l|c|c|c|l|c|c|c|}
\hline
          & $\alpha$ & $C$     & $p$ & & $\alpha$ & $C$     & $p$                      \\ \hline 
\multicolumn{4}{|c|}{Architecture} & \multicolumn{4}{c|}{Synthetic Noise} \\ \hline

Encoder-Decoder & 1.969 & 0.057 & \multirow{3}{*}{0.285} & No noise          & 1.969 & 0.064 & \multirow{3}{*}{0.296} \\
Decoder-only      & 1.817 & 0.11 & & Source noise   &   2.222 & 0.067 &                          \\
Hybrid-LSTM & 2.011 & 0.078 & & Target noise      & 2.772 & 0.323 &                        \\ \hline

\multicolumn{4}{|c|}{Filtering} & \multicolumn{4}{c|}{Back-Translation} \\ \hline
No filter & 2.501 & 0.034 & \multirow{3}{*}{0.278} & BT model 2L6L  & 2.343 & 0.059 & \multirow{4}{*}{0.198} \\
CDS       & 2.235 & 0.054 & & BT model 6L6L  & 2.288 & 0.054 &\\
Bicleaner & 2.130 & 0.064  & & BT model 32L6L & 2.251 & 0.040 &\\
& & & & BT model 64L6L & 2.224 & 0.037 & \\
& & & & Parallel data  & 1.196 & 0.048 &  0.271 \\
\hline

\end{tabular}
\end{sc}
\end{small}
\end{center}
\label{table:all-coeffs}
\end{table*}

\section{The Effect of Noise}
\label{sec:noise}
Large scale parallel corpora are essential to building high-quality machine translation models \citep{arivazhagan2019massively,lepikhin2020gshard,fan2020englishcentric}. Such datasets are created by crawling web pages and performing post-processing steps such as document alignment and sentence alignment \citep{elkishky2020ccaligned,banon-etal-2020-paracrawl} to create parallel data. However, there are many ways in which noise can enter this pipeline --- misaligned sentences, copied URLs, typos, mistranslations and so on. Such noise can potentially be detrimental to translation quality. For instance, prior work \citep{Khayrallah_2018} has found that adding a large amount of noisy data to high quality parallel data can have a catastrophic effect on the performance of NMT models. However, most such studies are performed on a fixed training dataset size.

In this section, we will study the effect of noise, not just by comparing noisy and clean data at a single dataset size, but by comparing their scaling for increasing dataset sizes. This allows us to ascertain if the impact of noise can be compensated with more data. We approach this question in two complementary ways (1) We start with a noisy web-crawled corpus (ParaCrawl English to German \citet{banon-etal-2020-paracrawl}) and apply filtering algorithms to it (Section \ref{sec:filtering}) (2) We start with a clean parallel corpus and add different types of noise to it (Sections \ref{sec:synth-noise}, \ref{sec:backtranslation}). 


\begin{figure*}
    \centering
    \includegraphics[width=0.7\linewidth]{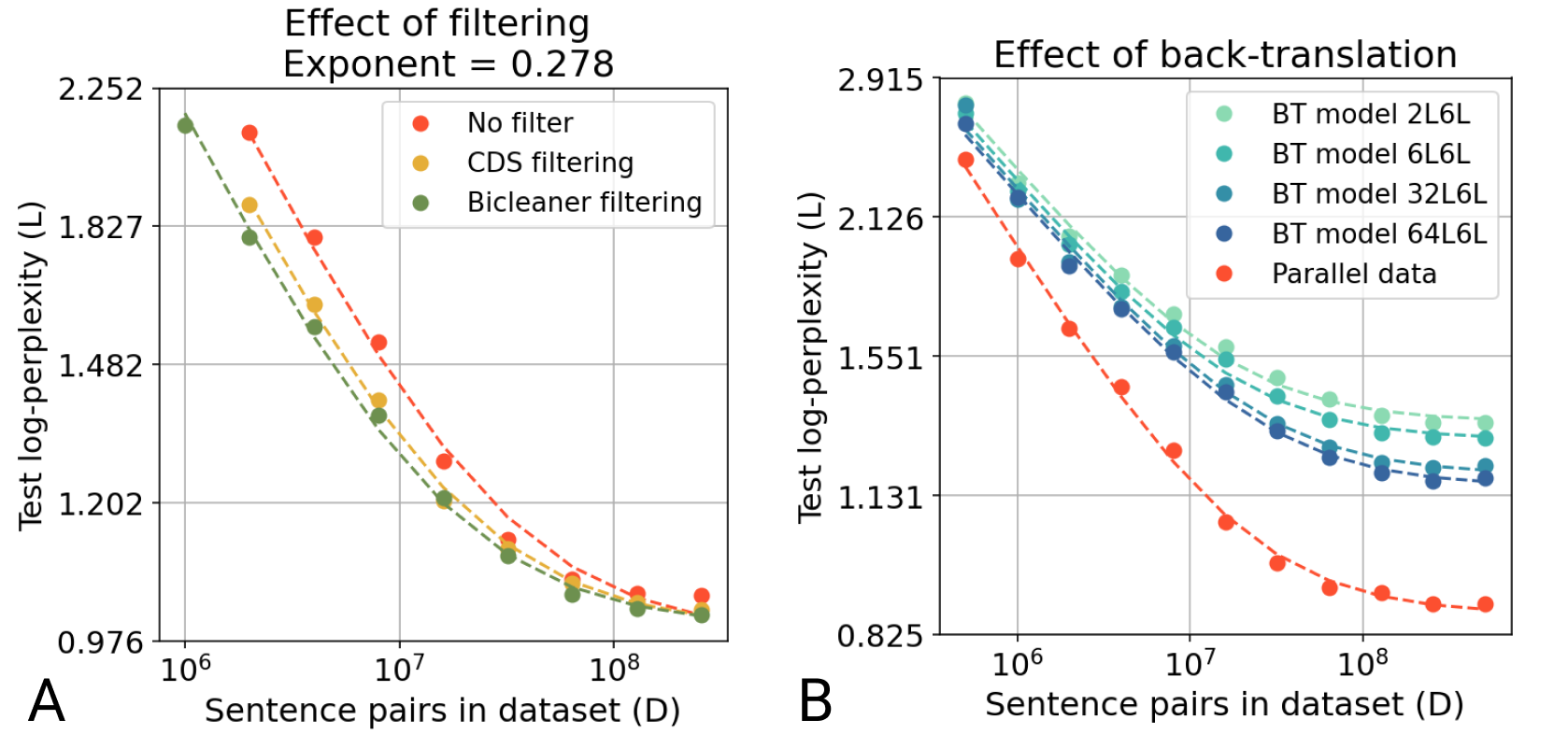}
    \caption{(A) {\bf Effect of filtering:} We apply two different filtering algorithms to the raw ParaCrawl dataset and evaluate the data scaling curve. We find that a common exponent provides a good fit for the experimental observations. (B) {\bf Effect of back-translation:} We train a 6L6L model on back-translated data from 4 different back-translation models \{2L6L, 6L6L, 32L6L, 64L6L\}. We find that the scaling exponent for back-translated data is worse than that for clean parallel data.}
    \label{fig:filter-bt}
\end{figure*}

\subsection{Data Filtering}
\label{sec:filtering}
Due to the prevalence of noise in web-crawled corpora and its impact on machine translation models, a variety of algorithms and heuristics have been developed to filter out noisy sentences \citep{wang2018denoising,junczys2018dual,ramirez-sanchez-etal-2020-bifixer}. Our goal here is to understand the impact of data filtering on data scaling.


{\bf Setup:} To understand the effect of data filtering, we use the largest publicly available ParaCrawl English-German dataset \citep{banon-etal-2020-paracrawl}. We lightly filter the raw dataset with de-duplication, length filtering ($\leq 256$) and language ID filtering. We also remove near duplicates of our test sets with a 10-gram overlap. This leaves $\sim 750M$ noisy sentence pairs. We train a 6L6L model on this dataset with increasing dataset sizes $\{1M, 2M...256M\}$. We compare two data filtering methods (1) Thresholding bicleaner scores \citep{ramirez-sanchez-etal-2020-bifixer} that are publicly released along with the ParaCrawl dataset v8.1. The bicleaner scores (ranging from $0$ to $1$) include various hard-coded rules, language-model fluency scores, and scores from a classifier trained to detect mutual translations. We use a threshold of $0.5$ and discard all sentences below this threshold leaving $\sim 300M$ sentence pairs. (2) Contrastive Data Selection (CDS) \citep{wang2018denoising}, which belongs to a family of cross-entropy-based filtering algorithms~\citep{moore-lewis-2010-intelligent,junczys2018dual}. CDS scores the quality of each sentence pair according to the difference in cross entropy scores between two related translation models: a clean model that was fine-tuned on a trusted dataset, and a noisy one that was not. We choose the top 50\% of the CDS-ranked sentences.


{\bf Results:} The results for the data scaling law for the Web Domain 1 test set on all three training datasets are shown in Figure \ref{fig:filter-bt}A. We fit a scaling law with a common exponent $p$ and separate $\alpha, C$ for each training set. The fitted coefficients are in Table~\ref{table:all-coeffs}. We find that a common exponent fits the experimental data well. We make the following observations: \textit{Multiplicative constant $\alpha$ shift:} The filtered dataset has lower $\alpha$, implying that at a given dataset size, the loss for filtered data is lower. Thus, if you were constrained by compute to use a small fixed dataset size, it would be advisable to use a filtered dataset. \textit{Loss at convergence $\alpha C^p$ is the same:} In our experiments, we find that the three datasets converge to the same loss for a 6L6L model. This implies that at large dataset sizes, a 6L6L model is unable to distinguish the differences between a filtered and unfiltered dataset. 
\textit{Similar exponents $p$:} We find that a common exponent is sufficient to describe the experimental data. Thus, more noisy data can be used to obtain the same performance as a smaller cleaner dataset.

Since there is no standard, task-independent definition or measure of sentence quality or noise, any filtering algorithm  runs the risk of biasing the training dataset, say, towards the in-domain trusted dataset used for filtering. For example, a recent study by \citet{gao2021empirical} shows that very aggressive filtering can negatively impact the downstream performance of language models. Our results show that the risk of this bias can be avoided by adding more data --- while {\bf \emph{some amount of filtering may be desirable for computational efficiency, we can replace filtered data with more unfiltered data}}.

{\bf Note on changing data source:} The experiments in this section are conducted on the ParaCrawl dataset as compared to the in-house dataset in Section \ref{sec:basic-scaling}. Both these data sources have different crawling pipelines and different distributions. However, interestingly, we find that the data scaling law for both source has a scaling exponent of $\approx 0.28$ as shown in Table \ref{table:all-coeffs}. This surprising consistency is more evidence that the data scaling exponent is robust to distributional changes.

\subsection{Adding Noise}
While understanding the effect of filtering on data scaling curves is practically informative, filtering combines many different types of noise and heuristics together. To have finer control on the types of noise, we now \emph{add} synthetic noise to a clean dataset. 
In particular, we make two different types of distinctions. First, we add noise either only to the source side (the input sentences) or only to the target side (the output sentences). Second, we consider \emph{independent vs. dependent noise}. We define independent noise as noise that does not depend on the source/target sentence itself, such as changing a character to a random character or deleting a random word. By dependent noise, we mean that added noise depends on the sentence itself, for example, if the word `cat' is always mistranslated as `dog'. We believe that this is a natural distinction since the effect of the former type of noise can (at least information-theoretically) be reversed if we are provided with enough data. On the other hand, dependent noise can bias the distribution in more drastic, irreversible ways. 


\vspace{0.1in}
\subsubsection{Independent noise}
\label{sec:synth-noise}
{\bf Setup:} We add the following types of iid noise added to the source and target side separately: (1) \textit{Character level:} We perturb $p=0.1$ fraction of the characters in the sentences to random characters (alphanumeric + punctuation), (2) \textit{Word level:} We delete $p=0.15$ fraction of the words, and (3) \textit{Sentence level:} For $p=0.1$ fraction of the sentences, we shuffle the mapping sentences of the sentence pairs. Thus, the source and target sentences have no correspondence. These noise types were studied previously by \cite{Khayrallah_2018}. Next, we train a 6L6L transformer model on increasing subsets of the noisy training datasets. The results are shown in Figure \ref{fig:intro}C and Table~\ref{table:all-coeffs}.

Our first observation is that we can fit Eqn. \eqref{eg:basic-scaling-law} with a common exponent $p$, but different $\alpha, C$ for the different training sets. On the other hand, unlike filtering, both the source and target noise datasets do not converge to the same loss value as the clean dataset at large dataset sizes $D \rightarrow \infty$. These results show that while the exponents for different datasets can be similar, it is also important to consider the loss where these models converge. In this particular case, more data cannot always offset the effect of noise. It is an open question if this is because the model size 6L6L is too small, if these types of noise are disruptive to neural network training irrespective of model size, or if this problem is hard to solve at finite samples for any class of models. 

Lastly, we find that target noise is more harmful to performance than source noise. This may help explain why backtranslation \citep{sennrich2016improving} is a useful data augmentation technique --- since it it uses a clean monolingual target corpus and noisy back-translated source sentences. Our results are in contrast with observations in the vision domain \cite{bahri2021explaining}, where changes to the input distribution change the exponent, but changes to the output distribution keep the exponent unchanged.


Given the strong emphasis on data quality in the NMT community, our experiments show that noise has less impact than one might have expected on the sample efficiency of NMT models. Both filtering natural noise and adding artificial independent noise have no impact on the exponent. However, they do impact the multiplicative constant, meaning that for a fixed computation or data budget, data quality remains quite relevant. Crucially, the techniques outlined here give practitioners tools they can use to help determine when effort should be put into removing noise, and when they should focus on collecting more data.

\subsubsection{Dependent noise: Back-translation}
\label{sec:backtranslation}

Now we turn our attention to changing the training distribution by training with back-translated data instead of parallel data. Back-translation (BT) \citep{sennrich2016improving} is a common data-augmentation technique employed in MT to increase the amount of training data. If you are training your model on English to German sentences, with back-translation, one would use a reverse model trained from German to English, and a clean monolingual German corpus to generate English-German sentence pairs. Back-translation can be considered a type of \emph{dependent noise} that is added to the source side. It differs from the independent noise considered in Section \ref{sec:noise} in that noise depends on the source/target sentence itself. For instance, if the BT model was never trained on any sentences on the topic of animals, it will make systematic errors on such sentences. This makes it an interesting setup to study data scaling, as it is not apriori obvious if such a distributional change would impact just the bias of the model $C$ or also the scaling exponent $p$.

{\bf Setup:} To minimize the number of confounders, we extract the German target side of the same dataset that was used to train our models in Section \ref{sec:basic-scaling}. This keeps the target distribution the same as the baseline. Then, we use four different German to English encoder-decoder models of sizes \{2L6L, 6L6L, 32L6L, 64L6L\} to generate English translations. This gives us four different datasets with English of varying quality, with the smallest model producing the `noisiest' source sentences. Note that these German to English models are trained on an different in-house dataset from the English to German dataset used in Section \ref{sec:basic-scaling}. However, the two datasets may contain some overlapping sentences. We examine the data scaling behavior for a 6L6L model trained on increasing random subsets of these datasets. 

{\bf Results:} The results of our experiments are shown in Figure \ref{fig:filter-bt}B. We fit a scaling law with common $p$ and dataset dependent $\alpha, C$. We find that the scaling exponent of the BT trained models is lower ($\sim 0.19$) than the scaling exponent of the parallel dataset ($\sim 0.28$) (See Figure \ref{fig:filter-bt}B). Moreover, the BT datasets converge to a worse loss value at the infinite data regime. As such, the utility of BT data is measurably lower than natural human-generated parallel data.

We find that increasing the BT model size does not affect the scaling exponent $p$. Instead, it improves the test performance by improving parameters $\alpha, C$ associated with the dataset. This improvement is especially pronounced in the capacity limited regime: In the data-limited regime, almost any back-translation model will provide good improvements, but as we approach very large dataset sizes, it will be more beneficial to use a larger BT model. Finally, the large gap between the parallel corpus and the best BT corpus at large data sizes show that even our largest BT models are far from fully capturing the complexities of human-generated translations.







\section{Conclusions}
We conducted a large-scale study of the changes in data scaling laws that occur with practically relevant changes to the training setup in NMT. We find that a majority of these changes lead only to a multiplicative shift in the scaling curves, and the exponent changes minimally. Thus, many advancements that seem significant at a small scale dataset, can be equivalently achieved by adding more data to move further across the data scaling curve. 


Apart from the practical implications, this work also raises interesting theoretical questions. If so many interventions to the training pipeline keep the exponents unchanged, then this may be indicative of a deeper commonality in the mechanism by which these deep networks learn. For instance, recent work \citep{bahri2021explaining, sharma2020neural} conjecture that the data scaling exponent captures the ``dimension of the data manifold'' as it is represented by the model. If this conjecture is true, then our experiments suggest that certain changes to the architecture or data distribution do not change this `manifold'. In all, the consistency of the data exponent across a variety of settings suggests that it captures a fundamental aspect of the learning problem.


\section{Acknowledgements}
We would like to thank Ankur Bapna, Ciprian Chelba, Markus Freitag, Xavier Garcia and Wolfgang Macherey for many insightful discussions. We would also like to thank Yasaman Bahri, Ethan Dyer and Yuan Cao for comments on an earlier version of this paper.

\bibliography{references}
\bibliographystyle{icml2022}

\newpage
\appendix
\onecolumn
\section{Scaling Law Fitting Details}

\subsection{Separate Fits and Variance}
\label{app:separate-fits}
\begin{figure}[h]
    \centering
    \includegraphics[width=\linewidth]{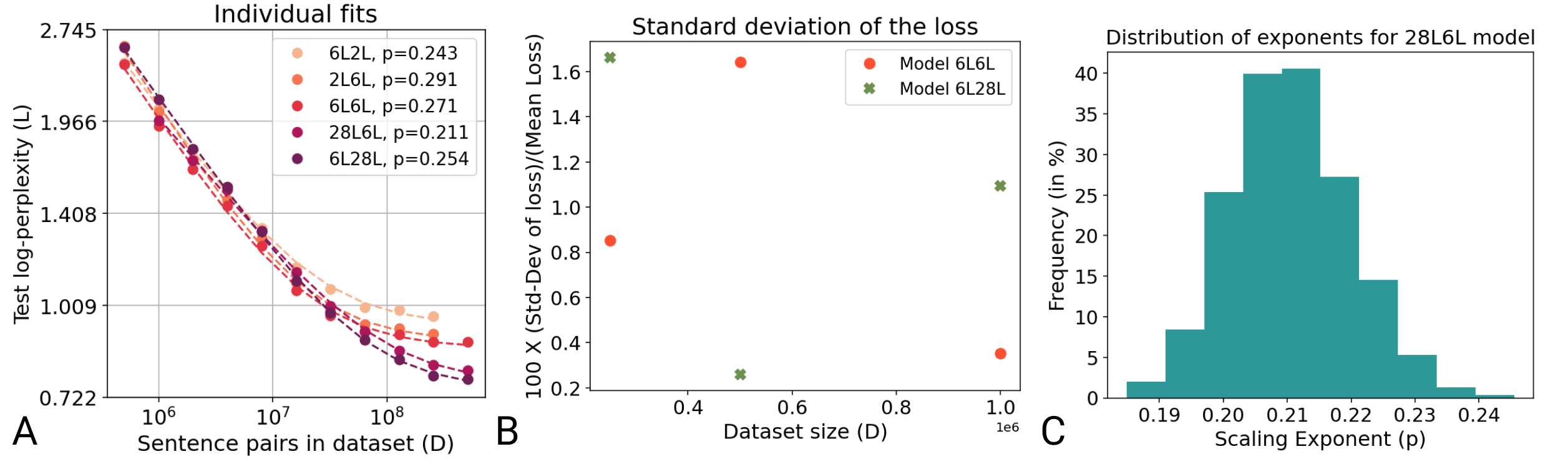}
    \caption{(A) Separate power law fits from Equation \ref{eg:basic-scaling-law} to different model sizes. We observe small differences in the scaling coefficients (B) Standard deviation in the loss due to randomness in training. The loss varies by up to 2\% (C) Distribution of the scaling exponents from a Monte Carlo simulation assuming a 2\% standard deviation in the loss.}
    \label{fig:variance}
\end{figure}

In Section \ref{sec:basic-scaling}, we showed that a single common exponent $p$ gives a reasonable fits to the experimental observations for different model sizes. We now examine the difference in the coefficients for each of these models. To do so, we fit a separate power law from Equation \ref{eg:basic-scaling-law} to each model as shown in Figure \ref{fig:variance}A. While the scaling exponents have minor differences, they are in the same range. 

Some of these variations can be attributed to the sensitivity of these scaling parameters to randomness in the training procedure. There are multiple sources of randomness in training --- initial random seed, randomness over sampling of the training set, randomness in SGD training such as batch order. To understand the effect of these, we sample 5 different versions of the training set for dataset sizes $\{250K, 500K, 1M \}$ (we choose these as they require low compute to train to convergence) and train networks on these datasets from scratch. Figure \ref{fig:variance}B shows the standard deviation observed in the test loss for the 6L6L and 6L28L models. As we can see, the variance is up to 2\% of the loss. The variance in larger datasets is expected to be lower than those for smaller datasets.

To understand how this variance would affect the final observed scaling coefficients, we do the following Monte Carlo simulation: We assume that the loss is distributed as $\mathcal{N}(l, 0.02l)$ where $l$ is the loss for a given dataset size and model. We then simulate different loss values from this distribution for all the dataset sizes, and fit scaling law from Equation \ref{eg:basic-scaling-law} to it. This gives us a distribution over the scaling coefficients. Figure \ref{fig:variance}C shows the distribution of the scaling exponents obtained from this procedure for the 28L6L model. As we can see, a 2\% randomness in the test loss, gives us a standard deviation of 0.02 in the scaling exponent. This provides a benchmark in comparing exponents obtained from two different experiments (say two different architectures). 

\subsection{Optimizing Hyperparameters}
\label{app:hyperparams}
In this section, we discuss the choice of hyperparameters in our experiments. We tune the learning rate such that the training loss is optimized optimally. We find that logit clipping, while important for stable training, only affects the final test loss minimally. On the other hand, dropout has a significant effect on the test loss. 

To understand how dropout affects the scaling law, we try a grid of dropout values for a range of number of samples for the 6L6L model. We find that the dropout value mostly affects test loss when the dataset size is less than $16M$. We construct the data scaling law for various dropout values, and also find the data scaling law for the Pareto optimal curve i.e. choosing the best dropout rate for each dataset size. We find that while the exact value of the test loss changes, the change in scaling exponent is relatively minor (from $0.27$ to $0.23$ for English to German). Thus, in our experiments, we use a dropout rate of $0.1$ across all settings, instead of fine-tuning the dropout rate for each setting separately (which is computationally prohibitive).

\begin{figure}
    \centering
    \includegraphics[width=0.6\linewidth]{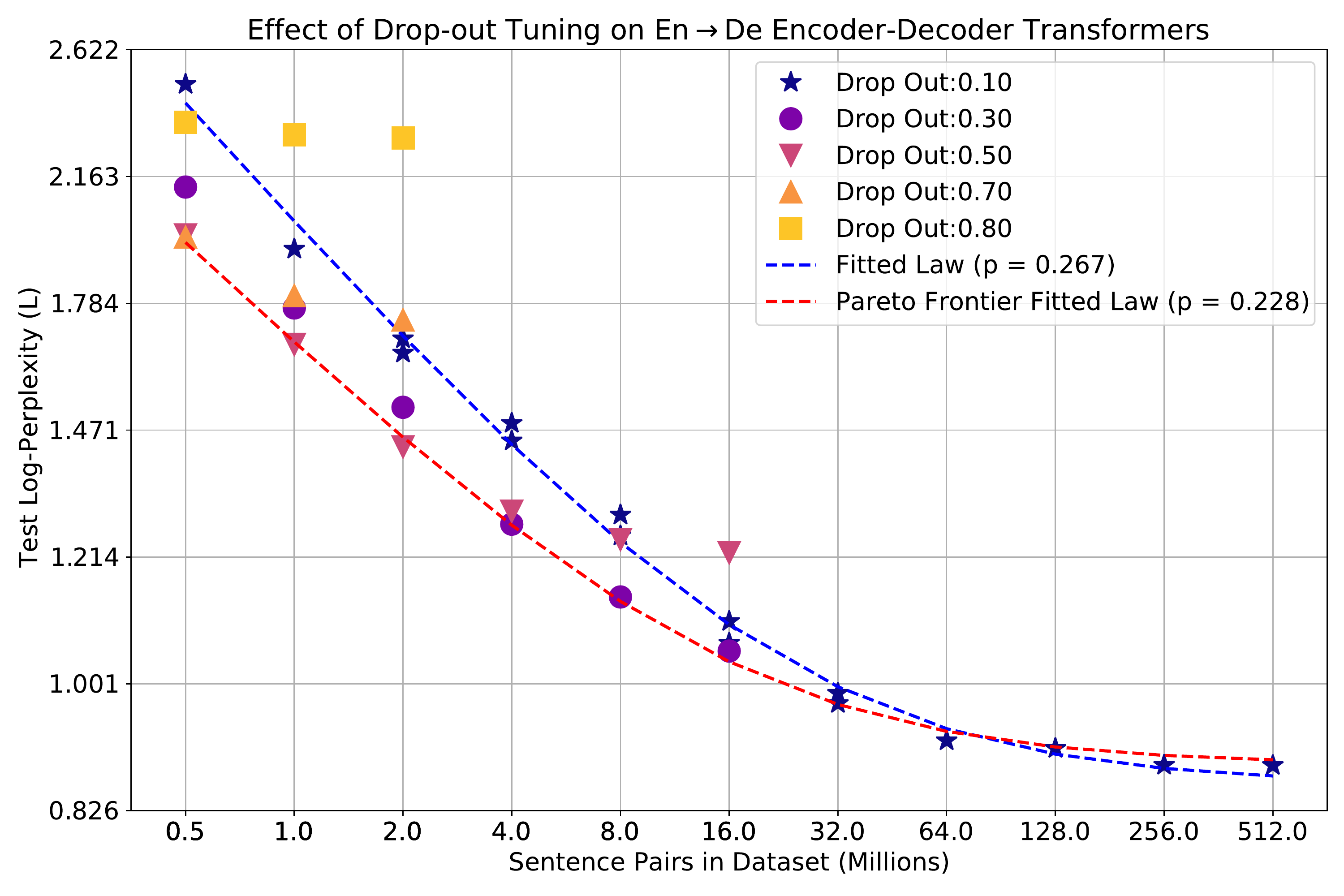}
    \includegraphics[width=0.6\linewidth]{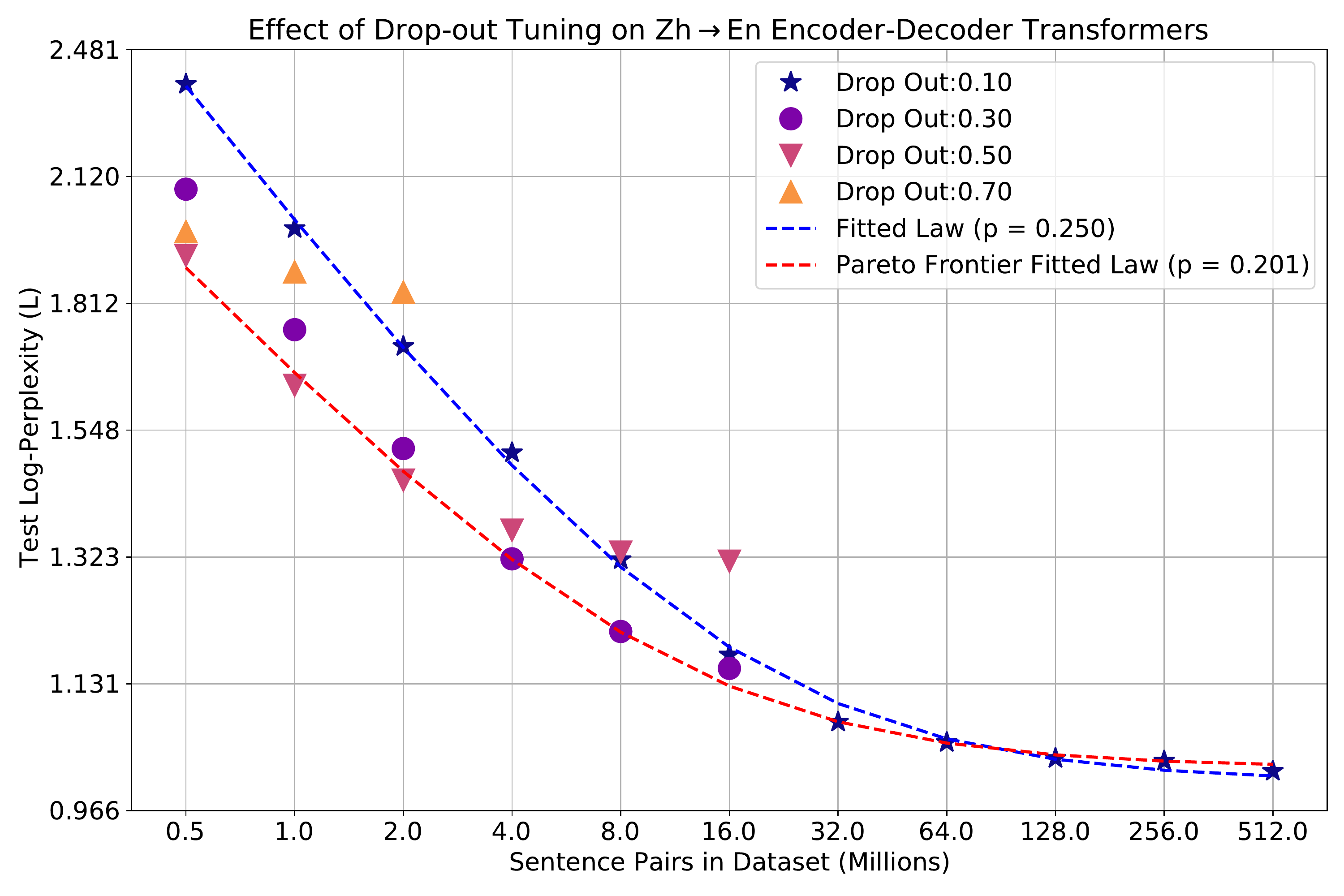}
    \caption{{\bf Effect of dropout on the scaling law.} We train a series of 6L6L transformer encoder-decoder models on English to German (top) and Chinese to English (bottom) translation tasks for a grid of dropout values. We find that the Pareto frontier and the single dropout rate data scaling curves have very similar exponents.}
    \label{fig:dropout}
\end{figure}


\subsection{Variance-Limited Regime}
\begin{figure}[h]
    \centering
    \includegraphics[width=0.4\linewidth]{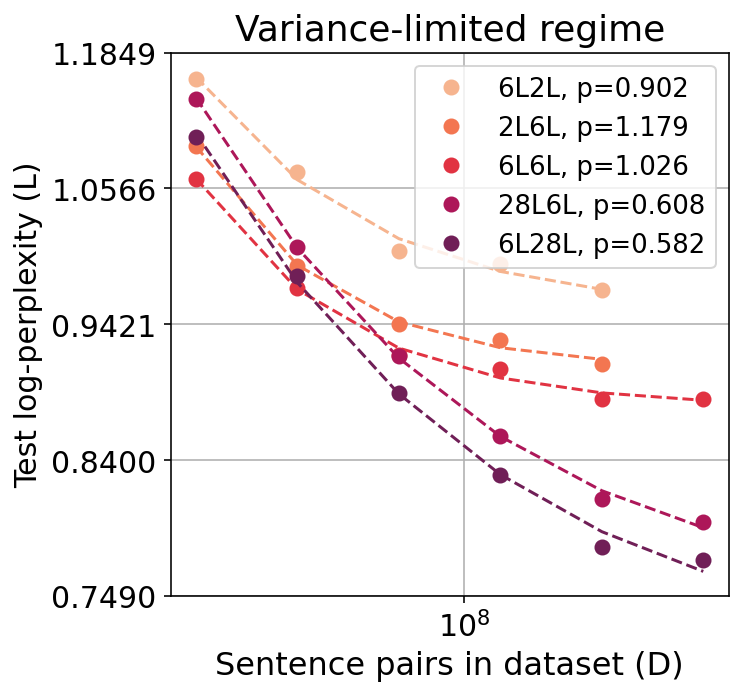}
    \caption{Power law fits for large dataset sizes show that the loss decays as $O(1/D)$}
    \label{fig:var-limited}
\end{figure}

We fit the scaling law $ L = \gamma(1/D)^p + B$ to dataset sizes $>= 32M$. If the ``variance-limited" conjecture is correct, then the scaling exponent should be $p \sim 1$. Figure \ref{fig:var-limited} indeed shows that this is the case. Moreover, the transition point to this regime occurs later for larger model sizes, as is expected from the form of the scaling law. As we can see, the exponents for the larger models has not reached $1$.

\section{BLEU Score Behavior}
\label{app:bleu}
Language tasks can roughly be categorized into two groups \textit{understanding} tasks where a given piece of text is tasked to be encoded for downstream classification (eg. sentiment analysis, named entity recognition), and \textit{generation} tasks where a representation of a piece of text is used (conditioned) to generate another arbitrary length sequence of text (e.g. summarization, question answering). Machine translation, without loss of generality, belongs to both of the categories: source content first needs to be understood and the corresponding target sequence must be generated. Given this categorization, we not only care about the model score on reference target sequence (measured in log-perplexity) but we also care about the generation quality. Once sequences are generated from a sequence model we resort to automatic measures like BLEU score. BLEU score is a precision based metric that compares a reference translation with the generated hypothesis by the model and yields a score between 0 and 1, taking into account n-gram overlap between reference and hypothesis while compensating for the lack of recall with a brevity penalty.

While we report BLEU in addition to log-perplexity scores in our study, we would like to bring the recent findings on the deficiency of BLEU as an automatic metric to the readers attention. As the MT systems have improved over the years, BLEU scores (along with several other automatic metrics) started to lose their sensitivity to approximate human judgement \citep{zhang2019effect,mathur-etal-2020-results,freitag-etal-2020-bleu,freitag2021experts,kocmi2021ship} and the translation community has started to experiment with learned metrics such as COMET \citep{rei-etal-2020-comet} or BLEURT \citep{sellam-etal-2020-bleurt}. Similarly, the methodology for performing human evaluations also remains an active and contested research area \cite{freitag-etal-2020-bleu}, in addition to human evaluations being expensive to perform. While the discussions around careful evaluation of generation quality are important, due to the ongoing debate around these issues and budget constraints, we report the log-perplexity and BLEU scores.


\section{Scaling Laws for Different OOD Datasets}
In all our main experiments, we study the scaling laws for a heldout test set from the same distribution as the training set. But we are also interested in the performance of out-of-distribution test sets. As such, we evaluate the model performance on a variety of other test sets covering a diverse set of domains and composition styles. To understand how performance on such test sets scales with dataset size, we measure the performance on various other test sets as described in Section \ref{sec:ood}. 

\begin{figure}[h]
    \centering
    \includegraphics[width=\linewidth]{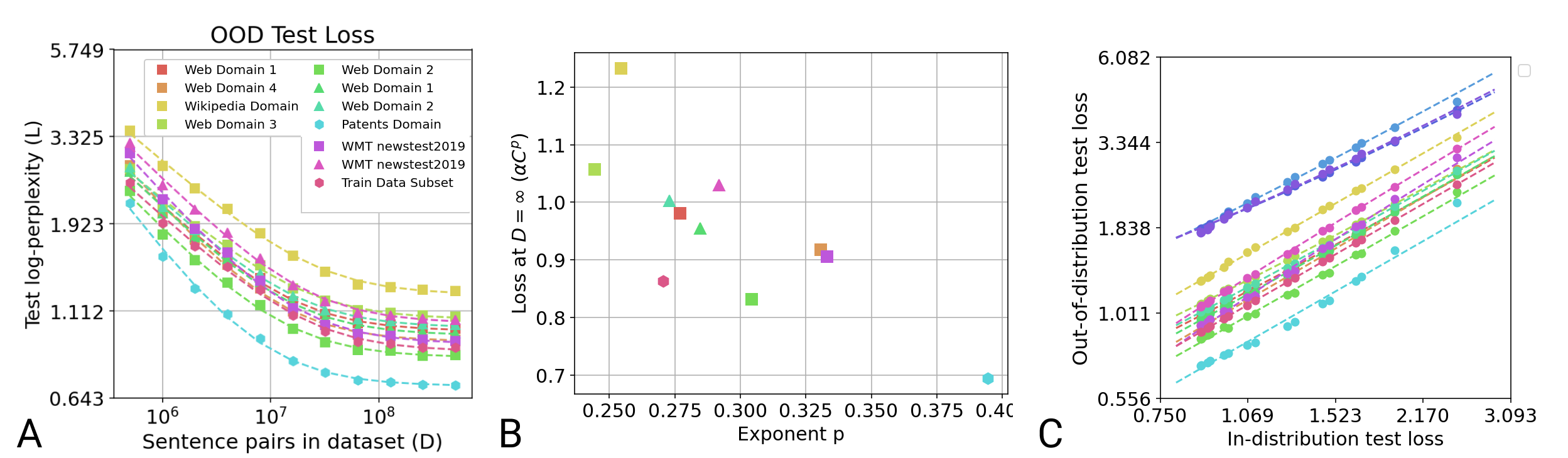}
    \caption{Scaling laws for various OOD test sets for 6L6L model}
    \label{fig:app-diff-tests}
\end{figure}

We find that the test loss on these additional test sets also follows a similar power law in the dataset size as Equation \ref{eg:basic-scaling-law}. Figure \ref{fig:app-diff-tests} shows the test loss fits for 14 different test sets for a 6L6L model. We find that most of the test sets have similar scaling exponents. That is, we do not observe any major differences on the basis on the test set composition --- both target and source original test sets scale similarly. 

Additionally, since the different test sets have similar scaling properties, this implies that the in-distribution loss and the out-of-distribution loss must have a nearly-linear relationship as we confirm in Figure \ref{fig:app-diff-tests}.

\section{Data Scaling Phase Transition} \label{app:phase-transition}
We fit the scaling law shown in Equation \ref{eg:basic-scaling-law}. 

\begin{equation}
    L = \alpha\Bigg(\frac{1}{D} + C\Bigg)^p
    \label{eg:basic-scaling-law-app}
\end{equation}

This equation displays two scaling regimes:
\begin{enumerate}
    \item Over-parameterized (or small $D$): In this regime, the $1/D$ term dominates the loss and the loss scales as $O(1/D^p)$. 
    \item Under-parameterized (or large $D$): In this regime, we can take a Taylor's approximation for the small term $1/D$ which leads to the loss scaling as $O(1/D)$.
\end{enumerate}

\begin{figure}[H]
    \centering
    \includegraphics[width=0.5\linewidth]{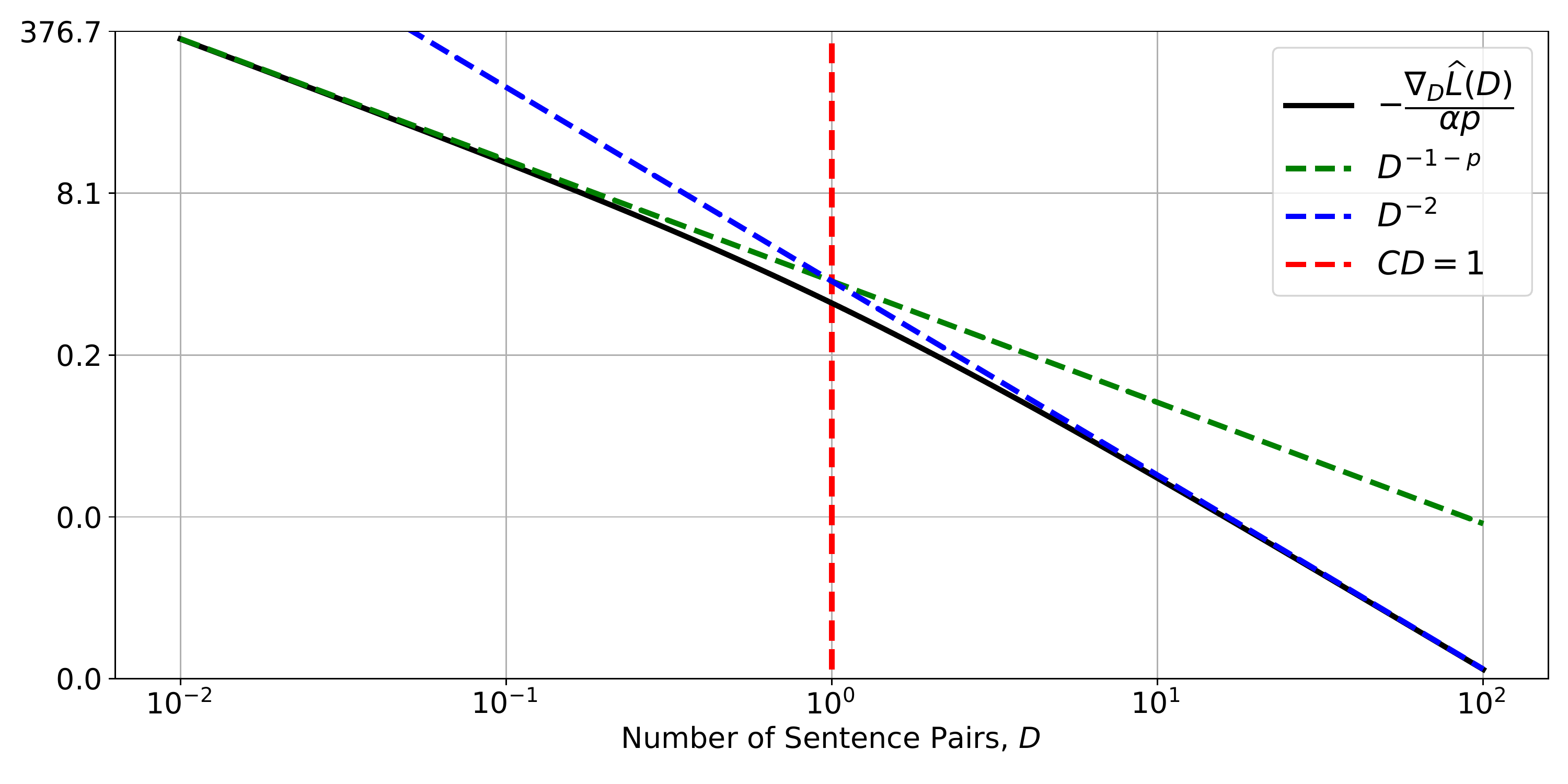}
    \caption{Phase transition from data-limited regime to model-limited regime.}
    \label{fig:phase-transition}
\end{figure}

Equating the derivatives of the two expressions provides an expression for the point where the marginal value of data transitions (and hence the model moves to capacity limited regime). A simple calculation shows that this point occurs at $CD=1$. Figure \ref{fig:phase-transition} provides an illustration for this transition.


\section{Scaling Laws with Different Architectures}
\label{app:architecture}




We now show additional plots for decoder-only and transformer-LSTM hybrid models, with individual fits for each model.

\begin{figure}[h]
    \centering
    \includegraphics[width=0.7\linewidth]{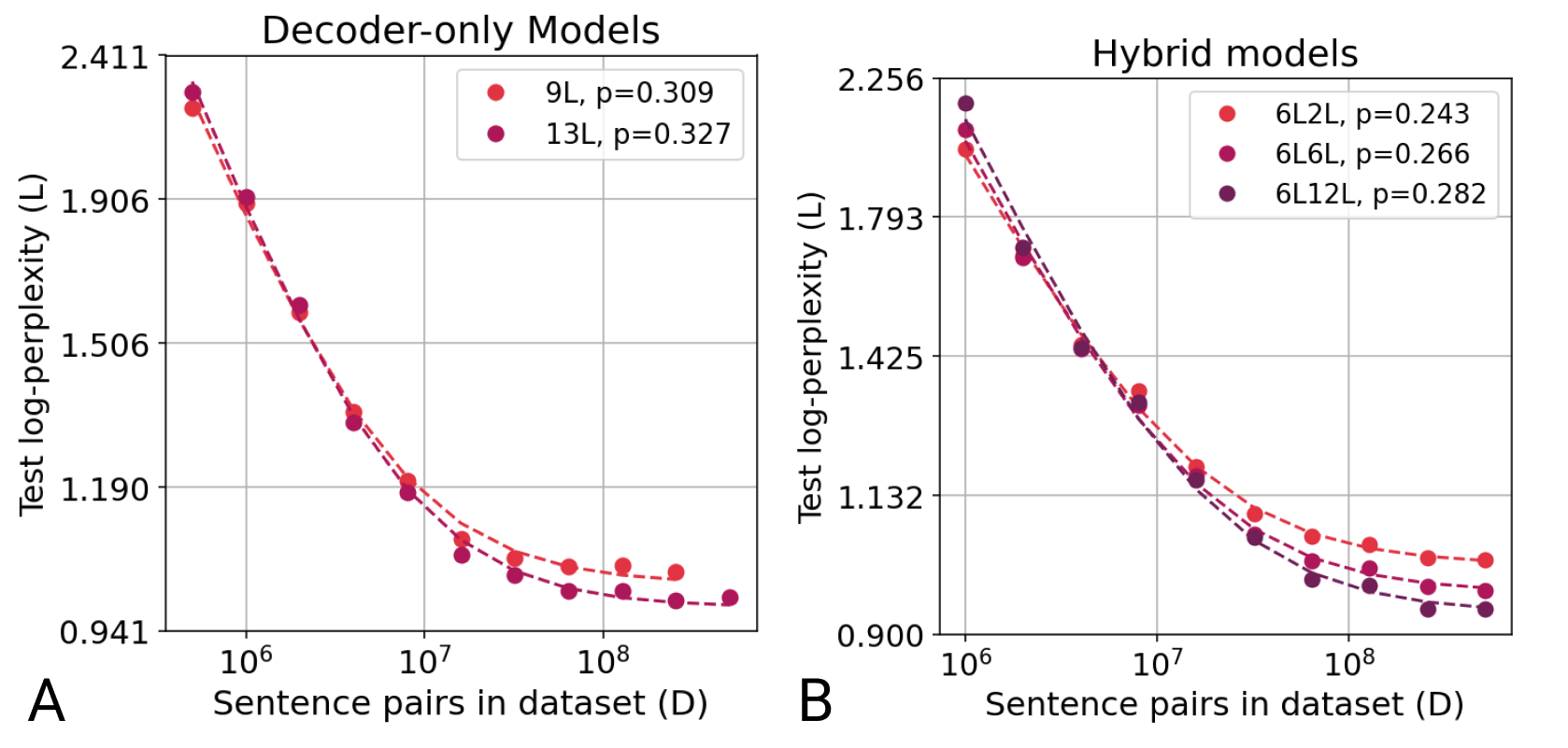}
    \caption{Separate fits for architecture with different depths (A) Decoder only (B) Transformer-LSTM hybrids}
    \label{fig:arch-more}
\end{figure}

\section{Changing Language Pairs}
\label{app:diff-lang}
To verify that our results are independent of the choice of language pair, we repeat a subset of our experiments for Chinese to English translation. Note that repeating all of our experiments for an additional language pair is computationally infeasible. Thus, we only repeat the experiments to compare different architectures.

{\bf Setup:} We use an in-house training dataset consisting of paired Chinese (source) and English (target) sentences. The training data pre-processing steps are the same as those described for the English to German training dataset. We train a 6L6L encoder-decoder transformer model, and a 6L6L transformer-LSTM hybrid. Additionally, we try a grid of different dropout values, ranging from $0.1$ to $0.5$. We take the `Pareto-frontier' i.e. the best dropout value for each dataset size and fit a data scaling law with a common exponent $p$ and different $\alpha, C$ for the two different model architectures.

{\bf Results:} Similar to English $\rightarrow$ German, we find that a common exponent is sufficient to describe the experimental observations.

\begin{figure}[h]
    \centering
    \includegraphics[width=0.5\textwidth]{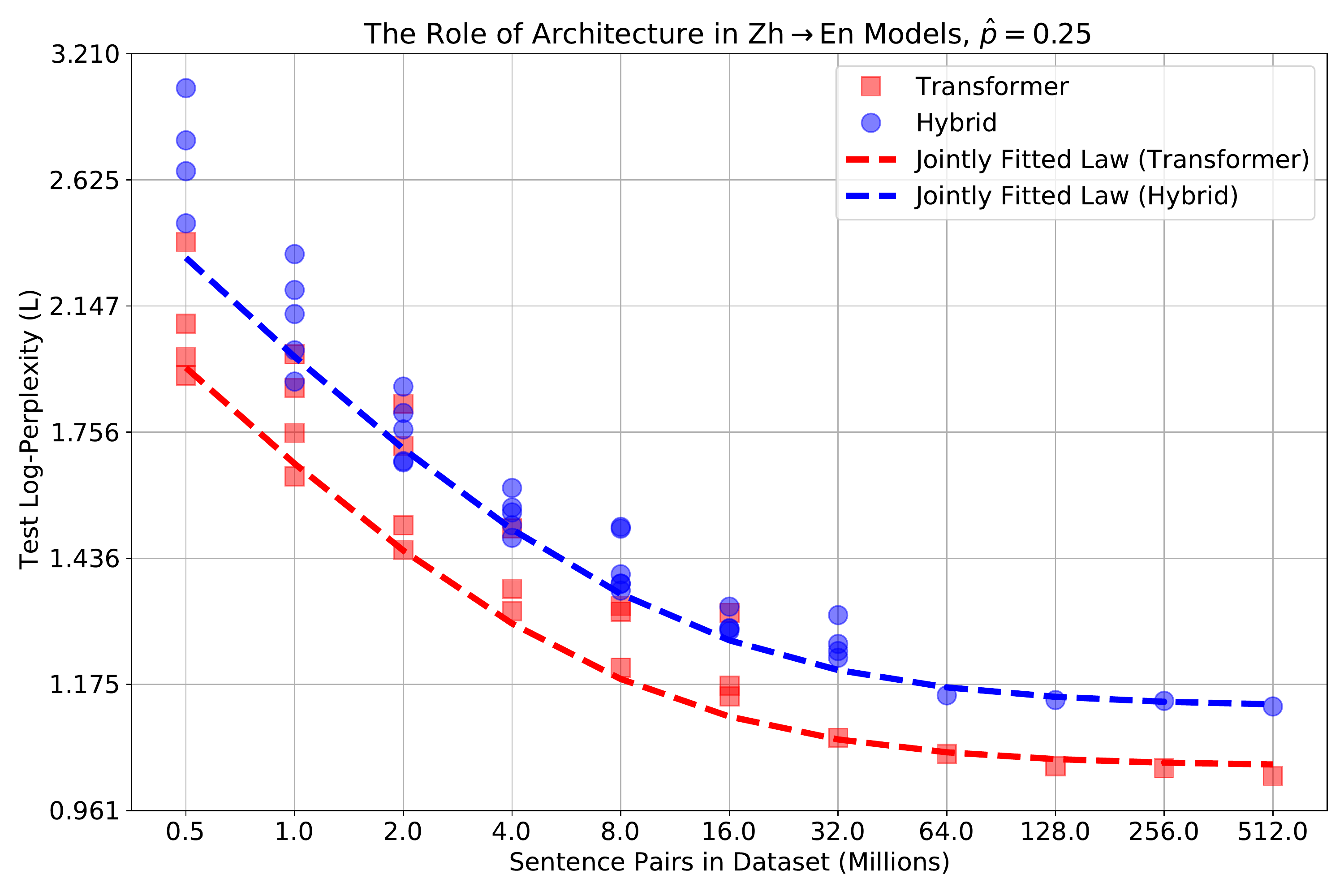}
    \caption{Architecture comparison for Chinese $\rightarrow$ English translation for an encoder-decoder transformer as well as a transformer-LSTM hybrid. The dotted line markers indicate the observations for different dropout values. The dotted line indicates the `Pareto-frontier' i.e. the best dropout value for each dataset size.}
    \label{fig:lang-arch}
\end{figure}

\begin{figure}[h]
    \centering
    \includegraphics[width=0.7\textwidth]{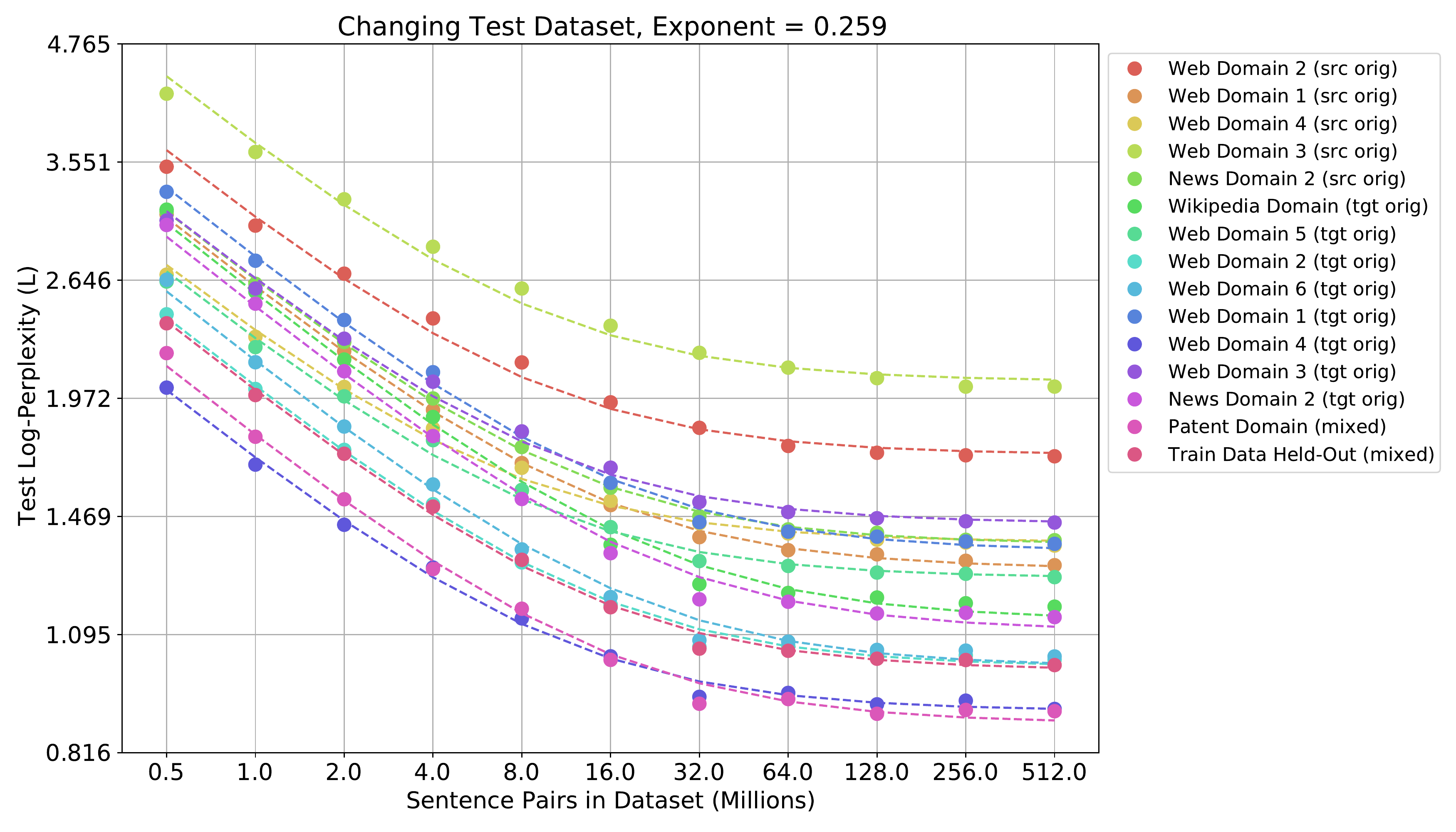}
    \caption{Out-Data scaling data scaling laws for Chinese $\rightarrow$ English translation for a 6L6L encoder-decoder transformer for a dropout rate $0.1$.}
    \label{fig:lang-ood}
\end{figure}




\end{document}